\documentclass[11pt,a4paper]{article}
\usepackage[hyperref]{conll-2019}
\usepackage{latexsym}
\usepackage{graphicx}
\usepackage{comment}
\usepackage{color}
\usepackage{paralist}
\usepackage{xspace}
\usepackage{amsmath}
\usepackage{amssymb}
\usepackage{enumitem}
\usepackage{array,multirow}
\usepackage{rotating}
\usepackage{pgfplots}
\pgfplotsset{compat=newest}
\usepackage{times}
\usepackage{url}
\usepackage{soul}
\usepackage{dblfloatfix}

\aclfinalcopy 

\usepackage{color, colortbl}
\definecolor{Gray}{gray}{0.925}
\newcolumntype{g}{>{\columncolor{Gray}}c}
\usepackage{booktabs}

\newcommand{\specialcell}[2][l]{\begin{tabular}[#1]{@{}l@{}}#2\end{tabular}}
\newcommand{\STAB}[1]{\begin{tabular}{@{}c@{}}#1\end{tabular}}

\usepackage{pifont}
\newcommand{\cmark}{\ding{51}}%
\newcommand{\Table}[1]{Table~\ref{#1}}

\newcommand{\Section}[1]{Section~\ref{#1}}
\newcommand{\Sections}[2]{Sections~\ref{#1} and~\ref{#2}}

\newcommand{\Figure}[1]{Figure~\ref{#1}}

\title{Say \textit{Anything}: Automatic Semantic Infelicity Detection \\ in L2 English Indefinite Pronouns}

\author{
	Ella Rabinovich$^{1}$\quad
	Julia Watson$^{1}$ \quad
	Barend Beekhuizen$^{2}$ \quad
	Suzanne Stevenson$^{1}$ 
	\vspace{0.1cm} \\
	$^{1}$Dept. of Computer Science, University of Toronto, Canada \\
	$^{2}$Dept. of Language Studies,\space\space University of Toronto, Canada \\
	\texttt{\{ella,jwatson,suzanne\}@cs.toronto.edu} \\
	\texttt{barendbeekhuizen@utoronto.ca}
}

\date{}

\begin{document}
\maketitle
\begin{abstract}
Computational research on error detection in second language speakers has mainly addressed clear grammatical anomalies typical to learners at the beginner-to-intermediate level. 
We focus instead on acquisition of subtle semantic nuances of English indefinite pronouns by non-native speakers at varying levels of proficiency.
We first lay out theoretical, linguistically motivated hypotheses, and supporting empirical evidence on the nature of the challenges posed by indefinite pronouns to English learners. We then suggest and evaluate an automatic approach for detection of atypical usage patterns, demonstrating that deep learning architectures are promising for this task involving nuanced semantic anomalies.

\end{abstract}

\section{Introduction}
The ubiquity of English as an online lingua franca offers a rich opportunity for computational research on second language acquisition and on tools for aiding non-native speakers.
%
Most computational research in second language (L2) has focused  on
spelling and grammar errors,
and has been conducted on learners with beginner-to-intermediate proficiency level  (henceforth, ``learners'')  
\cite[e.g.][]{ji2017nested, sakaguchi2017grammatical, rozovskaya2017adapting, lo2018cool}.  Little empirical
work has looked at
\textit{semantic} errors, with existing research mostly focusing on collocations \citep[e.g.,][]{dahlmeier2011correcting, vecchi2011linear, kochmar2013capturing}.  Also, highly proficient, advanced L2 speakers (henceforth, ``advanced L2s'') have received little attention \citep[though see][]{daudaravicius2016report}.
In contrast to learners, these speakers rarely violate grammatical norms of the L2, but rather 
deviate from native usage in much more nuanced ways, often exhibiting mild infelicities rather than outright errors.

\begin{table*}[hbt]
\centering
\resizebox{16cm}{!}{
\begin{tabular}{l|c|c|l}
\multicolumn{1}{l|}{Usage class} & 
\multicolumn{1}{c|}{\textit{some}-?} & 
\multicolumn{1}{c|}{\textit{any}-?} & 
\multicolumn{1}{l}{Example} \\
\hline
specific (SP) & \cmark &  & I had to reevaluate things when \textbf{someone} pointed that out.\\
non-specific (NS) & \cmark &  & {\textbf{Someone} please make me a GIF of that Wade dunk.}\\
question (QU) & \cmark & \cmark & {\textbf{Anyone} know what the issue might be?}\\
conditional (CD) & \cmark & \cmark & {I would love it if \textbf{someone} could explain it in a more precise way. }\\
indirect negation (IN) & \cmark & \cmark & {I don't understand how \textbf{anyone} can really hate on him.}\\
direct negation (DN) & \cmark & \cmark & {I don't have \textbf{anything} to add other than to say thanks for typing this out. }\\
comparison (CP) & \cmark & \cmark & {If you work harder you deserve to earn more than \textbf{someone} who doesn't do so.}\\
free choice (FC) &  & \cmark & {...they invite \textbf{anyone} on, including musicians sometimes.}\\
\end{tabular}
}
\caption{\small{Usage classes of IPs, an indication of those subsumed by \textit{some}- and \textit{any}-, and examples from our corpora.}}
\label{tab:functions-ips}
\end{table*}

We aim to explore an elusive aspect of mastering the subtle contours of a word's meaning that are shaped by its context. Specifically, we investigate patterns of acquisition of English indefinite pronouns by L2 speakers. \textit{Indefinite pronouns} (IPs) are linguistic devices that refer to an entity (such as a person or thing) that has not yet been introduced in discourse. In English, examples are words like \textit{someone}, \textit{anything}, and \textit{nobody}. Consider the following sentences, taken verbatim from corpora of L2 speakers (original pronoun is boldfaced; less felicitous usages marked with `?').\footnote{We refer to either less preferred or unacceptable occurrences of an IP, as in (2) and (3), as infelicitous usages.}

\vspace{-.05in}
\begin{enumerate}[leftmargin=*]
	\item \noindent \textit{Do you know \textbf{someone}/anyone who was discriminated based on gender?}
	\vspace{-.08in}
	\item \textit{It was a little amazing, because they didn't stole \textbf{?}\textbf{something}/anything.}
	\vspace{-.08in}
	\item \textit{\textbf{??}\textbf{Anyone}/Someone told me the company has millions in debts and isn't able to pay it.}
	\vspace{-.08in}
\end{enumerate}

\noindent
Clearly, mastery of IPs in English relies on recognizing subtle factors that determine their appropriate usage in various contexts. 

Here, in \Section{typ-view}, we develop a linguistic analysis with detailed hypotheses on precisely how the tangled relations between \textit{some-} and \textit{any-} pronouns, exemplified above, pose a challenge for L2 learners. In \Sections{data}{analysis}, we perform a large-scale investigation of these linguistic predictions using productions of both learners and advanced L2s, and find that the predicted infelicities occur not only in the language of the former but also the latter, albeit (as expected) to a lesser extent.

A practical goal of this work is to gain predictive power regarding the nuanced semantic difficulties that L2 speakers face. 
As a first step in that direction, in \Section{detection} we consider the ability of deep learning language models (LMs) -- shown to be adept at capturing grammatical phenomena \cite{ji2017nested, sakaguchi2017grammatical,marvin2018targeted, goldberg2019assessing} -- to identify the subtle infelicities that stem from the semantic confusion introduced by \textit{some-} and \textit{any-} IPs.  
We show that while state-of-the-art models obtain encouraging initial results on this task, they leave room for future improvement (possibly informed by our linguistic findings) in mastering the semantic nuances of the system of English IPs.

The contribution of this work is thus three-fold: First, to our knowledge, we develop the first large-scale empirical investigation of second-language acquisition of indefinite pronouns, constituting a case study of taking a computational approach in linguistic analysis to yield novel insights into challenges in L2 acquisition.
Second, we suggest and evaluate an automatic approach to detect infelicities stemming from these challenges in a large collection of L2 productions. Finally, in both cases, we extend our experiments to utterances of highly proficient L2 speakers -- a population that has heretofore received little attention in the context of automatic error/infelicity detection.\footnote{All code and data are available at \url{https://github.com/ellarabi/indefinite-pronouns}}

\section{Linguistic Insights into English IPs}
\label{typ-view}

Previous work has suggested that the English system of IPs is crosslinguistically atypical, with precise analogues to \textit{some-} and \textit{any-} unusual across languages \citep{haspelmath1997indefinite,beekhuizen2017semantic}.  Building on a suggestion from \citet{beekhuizen2017semantic}, we analyze the factors that could lead to difficulty in learning these IPs, and develop detailed hypotheses concerning the challenges that L2 speakers are predicted to face.

Our analysis is based on patterns of \textit{colexification} \cite{Francois2008}: that is, how usages expressing different semantics are grouped (or not) in various combinations under a single word.  
As the basis for our analysis,
we first need to specify the allowable semantic and syntactic usages of IPs.  These \textit{usage classes} are adapted from \citet{haspelmath1997indefinite}, who outlines a universal set of IP semantic functions across all languages.\footnote{Haspelmath's functions are determined by syntactic, semantic, and pragmatic factors.  Our usage classes emphasize the syntactic context, for ease of automatic identification and consistent annotation.}  Our usage classes are shown in \Table{tab:functions-ips}, with an indication of the classes that \textit{some}- and \textit{any}- can express.

\Table{tab:functions-ips} illustrates a striking fact about colexification of the usage classes in English: \textit{some}- and \textit{any}- each cover a very broad range of classes, with a high degree of overlap. This level of overlap in languages appears to be very rare: in the $40$ languages studied by \citet{haspelmath1997indefinite}, we find that only some $10$\% of languages have IPs that overlap over such a broad area of the semantic space.\footnote{Computed using the original Haspelmath's mapping into functions, therefore, not strictly comparable to the slightly different notation of usage classes in this work.}

Within any of these classes, some semantic/syntactic contexts call for just one of \textit{some}- or \textit{any}-, while others allow both, but with differing meanings (and frequencies/preferences).  For example, these similar contexts allow both, but the preferred pronoun differs:

\vspace{-.05in}
\begin{enumerate}[leftmargin=*]
\item \noindent \textit{...people care a lot \textbf{if something is a repost}...}
\vspace{-.08in}
\item \textit{...before you know \textbf{if anything is wrong}...}
\vspace{-.08in}
\end{enumerate}

\noindent
We thus predict a difficulty for English L2 speakers in having to choose between two (not interchangeable) terms that can be used in highly similar semantic/syntactic environments.

In addition to looking at difficulties posed by the colexification of IPs \textit{within English}, we can consider \textit{crosslinguistic} patterns of colexification for further insight.  Semantic typologists have proposed (and empirically supported, across many domains) that the more two underlying concepts are colexified across languages, the more similar those two concepts are \citep[e.g.,][]{anderson1982}.  In this way, crosslinguistic patterns of colexification can be used to deduce pairwise similarity among concepts, yielding a universal semantic similarity space for a domain \citep[e.g.,][]{berlin1969,levinson2003natural}.

\begin{figure}[]
\centering
\includegraphics[scale=0.65]{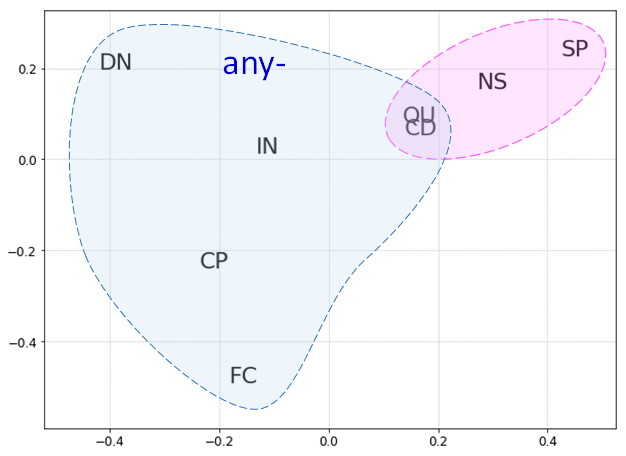}
\caption{\small{Layout of usage classes in crosslinguistic semantic space; light blue illustrates the scope of English \textit{any}-, pink illustrates the natural grouping of QU/CD with SP/NS.}
}
\label{MDS-plot}
\end{figure}

Here, we derive such a similarity space over the IP usage classes of \Table{tab:functions-ips}, using the colexification data across $40$ languages, from \citet{haspelmath1997indefinite}.\footnote{For this, we map our classes to Haspelmath's functions.}  We form a distance matrix (found in supplemental materials, A.1) by recording, for every pair of usage classes, the number of languages that have a term subsuming both those classes (indicating their relative similarity).  We then use Multidimensional Scaling (MDS) to project the space onto two dimensions, as exemplified in \Figure{MDS-plot}.\footnote{The relative distances slightly differ, but remain highly similar across many such projections we produce.}

\Figure{MDS-plot} demonstrates, first, that SP, FC, and DN form three natural ``extremes'' of the semantic space.  In English, these correspond to the canonical uses of the IPs \textit{some}-, \textit{any}-, and \textit{no}-, respectively; thus \textit{some}- is anchored at SP and \textit{any}- at FC (cf.\ \Table{tab:functions-ips}).
Moreover, we find that the usage classes of QU and CD are very close to SP and NS, indicating that QU and CD are most frequently colexified with SP/NS, in particular, much more so than with FC.
For English, this means that it is much more natural for \textit{some}- to express QU/CD than for \textit{any}- to do so.

To summarize, our linguistic analysis reveals two potential challenges of English \textit{some}- and \textit{any}-: their confusability across many classes, and the particular difficulty of \textit{any}- in the QU/CD classes.
We further find empirically that \textit{some}- IPs are more frequent than \textit{any}- in native English text, suggesting that \textit{some}- will be easier for L2 speakers, and that they may overgeneralize it when faced with uncertainty of which pronoun to use.  Collectively, these findings motivate:

\paragraph{Hypothesis 1:} The unusually large and overlapping extents of \textit{some}- and \textit{any}- are expected to pose difficulty for L2 speakers; \textit{any}- is predicted to be especially difficult due to its lower frequency.

\paragraph{Hypothesis 2:} Due to greater naturalness of grouping QU and CD with other classes subsumed by \textit{some}-, we predict that QU and CD usages of \textit{any}- will be particularly difficult for L2 speakers.

\vspace{.05in}
In exploring each of these hypotheses, we look for evidence in two forms: overuse of \textit{some}- compared to native speakers, and more errors involving \textit{any}-.
We focus on the frequent semantic categories of \textit{people} and \textit{things}, specifically the set of IPs \textit{someone}, \textit{anyone}, \textit{something}, and \textit{anything}.\footnote{We excluded \textit{somebody}/\textit{anybody} as they are about $1/10$ the frequency of their \textit{-one} counterparts in our data.}

\section{Materials and Methods}
\label{data}

\subsection{Datasets}
We expect that mastery of IPs will depend on a speaker's command of English, and therefore consider language productions both of learners (largely beginner-to-intermediate), and of L2 speakers on \href{https://www.reddit.com/}{Reddit} (shown to be highly proficient, almost on par with Reddit natives; \citealt{rabinovich2018native}).
Our learner dataset comprises several sub-corpora: EFCAMDAT \citep{geertzen2013automatic}, TOEFL11 \citep{blanchard2013toefl11}, and the freely available part of the FCE corpus \citep{yannakoudakis2011new}. The advanced L2 dataset includes online posts by advanced non-native English speakers from the L2-Reddit corpus \citep[released by][and comprising utterances by native as well as highly-proficient non-native speakers, published on the Reddit platform]{rabinovich2018native}. 
We extended the L2-Reddit corpus (originally collected in 2017) with data published through September 2018; the final dataset includes over $320$M native and L2 English sentences. Table \ref{tbl:datasets} presents details of the two corpora.

\begin{table}[hbt]
\centering
\resizebox{\columnwidth}{!}{
\begin{tabular}{lccc}
Dataset & Sentences & Tokens & L1s \\ \hline
learners & 5.6M & 72M & \textgreater13 \\
advanced L2s (Reddit) & 177M & 2.4B & 51 \\
native (Reddit) & 146M & 2.1B & -- \\
\end{tabular}
}
\vspace{-0.05in}
\caption{Statistics on datasets.}
\label{tbl:datasets}
\vspace{-2mm}
\end{table}

\subsection{Classification of IP Usages}
Evaluating our hypotheses in \Section{typ-view} depends on assessing which usage class an utterance with a \textit{some}-/\textit{any}- pronoun belongs to, so we can compare patterns of usage and infelicities across classes.  In English, the IP usage classes are often associated with particular lexical or syntactic cues in the clause with the IP -- e.g., a negative adverb for DN (\textit{I~don't want anything from this collection.}), or a question mark for QU (\textit{Would you like to buy something online?}). This enabled us to develop a rule-based classifier (see supplemental materials (A.3) for details), using a parser \cite{kitaev2018multilingual} and a set of heuristic rules.

We evaluated the classifier on sentences manually annotated by three in-house native English speakers with a background in linguistics. A sample of $750$ sentences produced by Reddit native English speakers was selected for annotation, and the annotators assigned a label to each sentence from within the set of \{DN, QU, CD, CP, MIXED\}, where the MIXED class comprises the SP, NS, FC, and IN classes (cf.~\Table{tab:functions-ips}). The MIXED grouping contains classes that are (1) difficult to distinguish using simple lexical and syntactic cues (essentially, an ``other'' class), and (2) predicted by our linguistic analysis to be relatively similar in their error patterns. Average annotator agreement on our task was $\kappa = 0.932$; detailed annotation guidelines can be found in supplemental materials (A.2).

\Table{tbl:functions-classification} shows that our rule-based classification is a reliable way to categorize a sentence with an IP (five-way classification baseline is $0.2$). Because we use a subset of sentences associated with each usage class throughout our experiments, we focus on classification precision, while maintaining recall. We use this classifier to automatically label L2 sentences by usage class.

\begin{table}[hbt]
\centering
\resizebox{\columnwidth}{!}{
\begin{tabular}{l|rrrrr}
Class & DN & QU & CD & CP & MIXED \\ \hline
P & 0.835 & 0.882 & 0.853 & 0.833 & 0.849 \\
R & 0.723 & 0.789 & 0.853 & 0.962 & 0.874 \\
F1 & 0.775 & 0.833 & 0.853 & 0.893 & 0.861 \\
\end{tabular}
}
\caption{\small{Evaluation of classification of IP usage classes.}}
\label{tbl:functions-classification}
\end{table}

\subsection{Annotation of (In)felicitous Usages}
\label{annotation}
We used the \href{https://www.figure-eight.com/}{FigureEight} crowdsourcing platform for collecting annotations to be used as ground truth of L2 infelicities. We extracted a randomly sampled set of $3,711$ sentences from our learner corpus representing a balanced distribution over the five usage classes,\footnote{Aiming at $1$K per class, limited by 587 and 124 sentences in the QU and CP classes in our learner data, respectively.} and a similar set of $10,000$ sentences from our advanced L2 (Reddit) corpus, each containing a usage of \textit{someone}, \textit{something}, \textit{anyone}, or \textit{anything}.\footnote{We excluded sentences with idiomatic expressions containing IPs from this work; see supplemental materials (A.5).} Each sentence was annotated by five native English speakers in a choice-based annotation scheme. The occurrence of the IP in the sentence was replaced with a blank line, and each annotator marked their preference for the \textit{some-} or \textit{any-} pronoun in that context (or ``other''), reflecting the most natural choice between the two. The gold annotation for each sentence was determined by its majority choice, and the confidence score was computed based on the number of selections (out of five annotators) of each of the two pronouns. Annotation guidelines and a sample of 500 manually annotated sentences can be found in the supplemental materials (A.4).

Table~\ref{tbl:annotation-examples} presents example sentences produced by learners and L2 Reddit authors where the majority annotation unanimously differed from the original pronoun (as indicated). The utterances are provided verbatim, maintaining grammatical errors typical to productions in our corpora.

\begin{table*}[hbt]
\centering
\resizebox{16cm}{!}{
\begin{tabular}{ll}
L2 utterance & Annotation \\ \hline
{\specialcell{Moreover, he also takes a risk of not knowing \textbf{someone} from this country.}} & anyone \\ \hline
{\specialcell{About 20 years ago, we didn't know \textbf{someone} who cares about them, who defend animal's right, \\ but today, I know many people who cares about, cause animals need to be protected.}} & anyone \\ \hline
{\specialcell{It is justified to say that they have to change \textbf{anything} to cope with the now situation.}} & something \\
\hline \hline
{\specialcell{I never said \textbf{something} about political science, probably it was not very good worded but my point is \\ just that it shows how the extremes of two sides can come closer together again.}} & anything \\ \hline
{\specialcell{I think it's a sampling bias rather than \textbf{anyone} massaging the numbers to see what they want to see.}} & someone \\ \hline
{\specialcell{If there is a day where no one works then this is useless because you can't do \textbf{something} on that day \\ with family besides walking in forests because everything would be closed.}} & anything \\
\end{tabular}
}
\caption{\small{Example sentences annotated by human annotators for infelicitous pronoun choice (original pronoun is boldfaced). The top part refers to learners' utterances, the bottom part refers to advanced L2s'.}}
\label{tbl:annotation-examples}
\vspace{-2mm}
\end{table*}

Sentences with a confidence level $\leq0.6$ are considered close to equally felicitous with either pronoun, while the confidence of 1 represents a unanimous preference for one of the alternatives.  Because we used a forced-choice task, if both pronouns were acceptable (e.g., \textit{Did you see something/anything you like?}), we expect that the confidence score will indicate the level of naturalness or typicality of the pronoun in that context. For this reason, we only consider an example infelicitous when it differs from annotator choice with a confidence $\geq0.8$, which indicates a stronger preference for one pronoun over the other. 

The final annotation results include $50$\% ($1556$) and $77$\% ($2857$) of sentences with a confidence of $1.0$ and of 
$\geq0.8$, respectively, for learners. Our advanced L2 data has $56$\% ($5639$) of sentences with a confidence of $1.0$ and $81$\% ($8079$) of 
$\geq0.8$.

A question arises as to how meaningful it is to label an IP usage as infelicitous -- i.e., the preferred IP in annotation differed from the original -- 
if both \textit{some}- and \textit{any}- are in fact acceptable. To explore this, we also got crowdsourced annotations on $500$ \textit{native} utterances from Reddit, and compared the percentages of usages annotated as infelicitous to those of $500$ randomly sampled sentences by advanced L2s. We found that $3$\% of native utterances were annotated as infelicitous at a confidence level of $\ge 0.8$, indicating a high agreement among native writers and our annotators, while for advanced L2s, the percentage was around twice that high -- $6.7$\%. Despite acceptable variation in \textit{some}-/\textit{any}- usage in a given context, even advanced L2 speakers differ from natives in their relative preferences.


\section{Analysis of IP Infelicities in L2}
\label{analysis}

\subsection{Distribution of IPs by Usage Types}
\label{ip-dist}

First, considering \textbf{Hypothesis~1} from \Section{typ-view}, we expect the confusability of \textit{some}- and \textit{any}- to be reflected in overgeneralization of \textit{some}- due to its higher frequency.
The subtle distinction between these pronoun types is assumed to be better mastered by advanced L2 speakers, so we expect the divergence from the native distribution to be amplified in learners' productions.  

\begin{figure*}
\includegraphics[scale=0.87]{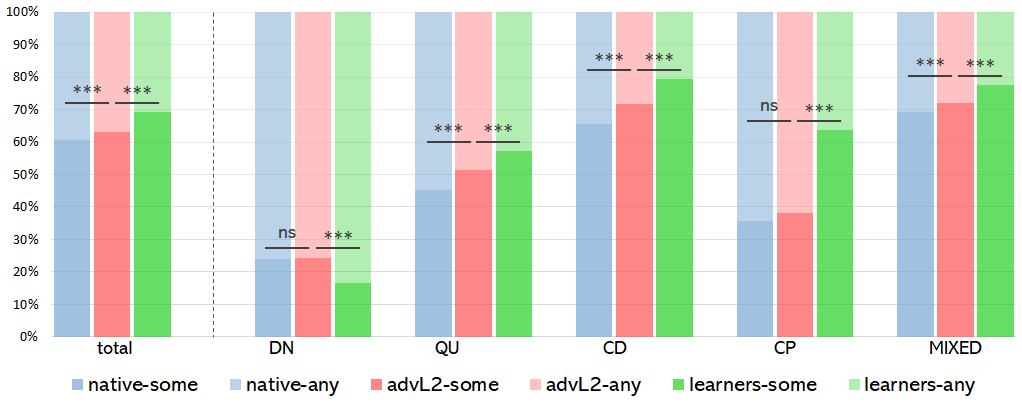}
\vspace{-0.02in}
\caption{\small{Distribution of \textit{some}- and \textit{any}- pronouns by usage class (native, advL2, learner, left-to-right in each); see \Table{tab:functions-ips} for definitions of classes. `total' refers to \textit{some}- and \textit{any}- counts extracted from the sample of $5$M sentences for each population. `***' indicates significant difference at the level of $p<.001$; `ns' indicates non-significant difference.}}
\label{fig:pronouns-by-function}
\end{figure*}

\Figure{fig:pronouns-by-function} presents relative frequencies of \textit{some}- and \textit{any}- pronouns in a random sample of $5$M native, advanced L2, and learner productions, both in the entire sample (left) and distributed by usage class (right). In line with our predictions, we find in Figure \ref{fig:pronouns-by-function} (left) that overall, L2 speakers use \textit{some}- pronouns more than \textit{any}- pronouns compared to native speakers.
We can further see in \Figure{fig:pronouns-by-function} (right), and discussed in detail below, that this pattern occurs in almost all the IP usage classes, especially pronounced for learners.

Elaborating on \textbf{Hypothesis 1}, we further suggest that
in addition to \textit{general} overuse of \textit{some}- vs. \textit{any}- (which may partly be due to avoidance of \mbox{\textit{any}-)}, L2 speakers are also expected \textit{in their infelicities} to more often use \textit{some}- where native speakers would use \textit{any}-, than vice versa. This prediction is also supported by our annotated data: In cases where the preferred pronoun is \textit{some}-, learners infelicitously use \textit{any}- $8.4$\% of the time, but in cases where the preferred pronoun is \mbox{\textit{any}-,} learners infelicitously use \textit{some}- almost $23$\% of the time. That is, learners have almost three times as many infelicities of using \textit{some}- instead of \textit{any}- than the reverse. Our advanced L2s speakers also show more infelicities using \textit{some}- instead of \textit{any}- than vice versa, but the difference is less pronounced ($5.8$\% and $10.1$\% respectively), as we expect given their greater proficiency.

\subsection{Distribution of Infelicitous Usages}
\label{class-dist}

Next we turn to \textbf{Hypothesis 2} from \Section{typ-view}, which further predicts that the precise extent of deviation from native-like usage patterns will not be distributed uniformly across the different usage classes, but rather there will be a higher degree of deviation in classes that are atypically grouped under \textit{any}- -- that is, QU and CD -- than in those that introduce less of a semantic challenge (DN, CP, and those in the MIXED class). L2 speakers are expected to exhibit both more overuse of \textit{some}- and more infelicities in the QU and CD classes.

Our predictions regarding the non-uniform overuse of \textit{some}- are largely borne out in Figure \ref{fig:pronouns-by-function}: the classes expected to be most difficult for L2 speakers -- QU and CD -- show a significant difference not only for learners, but even for advanced L2 speakers compared to natives, while DN and CP show only a difference for learners. 

A few observations from \Figure{fig:pronouns-by-function} do not follow our hypothesis.  First, the difference in learner usage of \textit{some}- vs. \textit{any}- for DN goes in the direction opposite to the prediction: i.e., learners use \textit{any}- more than \textit{some}- pronouns in direct negation.  We attribute this to the sheer frequency of \textit{any}- in direct negation, such that learners are overgeneralizing \textit{any}- here.  Second, the MIXED grouping also shows a difference for the advanced L2 speakers, although these usages are not predicted to be especially difficult by our linguistic analysis. This class contains a very large and diverse set of usages, making it difficult to predict what is driving this effect, and we leave this for future work. Finally, the largest gap in overuse of \textit{some}- vs. \textit{any}- is observed in the CP class for learners, thereby not complying with our prediction of the highest difficulty being introduced by the QU and CD classes. Note, however, that this result is based on a relatively small amount of data in the CP class for learners (only $124$ sentences; see Table \ref{tbl:infelicities-dist}).

To consider the pattern of infelicities across the usage classes, \Table{tbl:infelicities-dist} shows the results from our crowdsourced annotation of IP usages of learners (top) and advanced L2s (bottom), separated by the classes. As expected, learners exhibit a very high percentage of infelicities in the QU class ($24$\%); the CD class is not nearly as bad ($12$\%), but is still higher than the other three ($8$--$9$\%). Although advanced L2s have much fewer infelicities than learners, they also have more in the QU and CD classes ($7$\% and over $9$\% respectively) than in the others ($5$--$6$\%). Thus, as with Hypothesis 1, \textbf{Hypothesis 2} is largely borne out by the data, and we find additional evidence that the IP system of English is particularly challenging for beginning to intermediate learners.

\begin{table}[hbt]
\centering
\resizebox{\columnwidth}{!}{
\begin{tabular}{l|rggrr}
Usage class & DN & QU & CD & CP & MIXED \\ 
\midrule[0.01mm]
\# annotated & 1000 & 587 & 1000 & 124 & 1000 \\
\# infelicitous & 81 & 141 & 124 & 11 & 87 \\
\% infelicitous & 8.1 & 24.0 & 12.4 & 8.9 & 8.7 \\
\midrule[0.01mm]
\# annotated & 2000 & 2000 & 2000 & 2000 & 2000 \\
\# infelicitous & 106 & 141 & 182 & 102 & 113 \\
\% infelicitous & 5.3 & 7.1 & 9.1 & 5.1 & 5.7\\
\end{tabular}
}
\vspace{-0.03in}
\caption{{Distribution of annotated infelicities by usage class. Top panel: learners; bottom: advanced L2s.}}
\label{tbl:infelicities-dist}
\vspace{-2mm}
\end{table}

\vspace{-0.075in}
\section{Automatic Detection of Infelicities}
\label{detection}
Our motivation for the above analysis is to use these insights to drive development of tools for L2 learners. Here we consider the first step, that of detection of infelicities with a language model (LM). 

Neural network based approaches are currently among the most successful LMs. While being easily applied to a wide range of tasks, they provide significant improvements over classic backoff n-gram models. A common use of a pre-trained LM -- typically trained on an extremely large corpus -- is to predict the likelihood of an `unseen' sample of text: The higher the score (or the lower the perplexity) a text is assigned, the more probable it is, given the model. In particular, a fluent, well-formed text is likely to be scored higher by an LM than a text containing linguistic anomalies.

Encouraged by results on the task of grammatical error detection \cite{yuan2016grammatical, ji2017nested}, we adhere to a similar approach, casting the detection of infelicities as a binary classification scenario: An LM is applied on a sentence with an original pronoun (e.g., \textit{something}) and on the same sentence where the pronoun is substituted with its alternative (e.g., \textit{anything}); then the one predicted as more probable (scored highest) is chosen as a model decision. 

\subsection{Models}
Aiming to test the effect of various factors, such as training data size and register, on the predictive power of LMs in our task, we used both pre-trained models and models trained locally on in-domain, albeit much smaller, data.

\paragraph{\href{https://github.com/facebookresearch/colorlessgreenRNNs/tree/master/src/language_models}{Gulordava et al.}:} A successful variant of RNNs, the long short-term memory model \citep[LSTM,][]{hochreiter1997long}, used for syntactic error detection in \citet{gulordava2018colorless}. 
We trained the model using a similar set of parameters to \citet{gulordava2018colorless},\footnote{Specifically, we used two hidden layers of $200$ units per layer, dropout rate of $0.2$, batch size of $20$, and initial learning rate of $20$, and trained for $40$ epochs (until the validation set perplexity converged).} on 10M sentences by native English speakers of Reddit (see Section \ref{data}), using a 20K sentence validation set and a 50K sentence test set. This model allows us to test the benefits of using \textit{in-domain} data (for advanced L2s), despite its significantly lower volume, compared to other models.

\paragraph{\href{https://github.com/tensorflow/models/tree/master/research/lm_1b}{Google 1B}:} A very large publicly available LM released by \citet{jozefowicz2016exploring}. This fine-tuned language model, trained on a billion-word corpus \cite{chelba2013one}, requires a massive infrastructure for training. It achieves impressive perplexity scores on common benchmarks, and has been shown effective on a range of NLP tasks.

\paragraph{\href{https://github.com/huggingface/pytorch-pretrained-BERT}{BERT}:} A recent bidirectional encoder representations from transformers (BERT) LM released by Google \cite{devlin2018bert}. Proven highly effective in several language modeling tasks, it achieves state-of-the-art results in syntax-sensitive scenarios \cite{goldberg2019assessing}, pushing the limits of what is feasible with current language modeling tools.


\vspace{.1in}
We report the models' precision, recall and F1 scores for infelicitous and correct classes separately. We also report the overall accuracy of each, computed as the ratio of correctly classified cases out of all sentences. Following the intuition laid out in \Section{annotation}, we conducted two sets of experiments: (1)~considering cases where annotators' confidence score was $0.8$ or higher, and (2)~considering cases with confidence of $1$. Sentences with a lower confidence score  (i.e., where both \textit{some}- and \textit{any}- were roughly equally preferred) were excluded from these experiments.

\subsection{Results and discussion}

Tables~\ref{tbl:detection-learners} and \ref{tbl:detection-reddit} present the results for learners and advanced L2 speakers, each split by the degree of annotation confidence. Baseline accuracy is computed as the ratio of felicitous usages (the majority class) out of all instances.  The Gulordava et al.\ LM yields results inferior to the baseline, despite training on in-domain (but much smaller) data.  BERT performs best overall, and both it and Google 1B exceed the baseline for learners, but BERT performs only at baseline for advanced L2s, confirming the extreme difficulty of this task.
Results obtained for the correct class are far superior to those for the infelicitous class, suggestive of the inherent difficulty of the latter cases, compared to (occasionally clear-cut) correct usage patterns. 
\begin{table*}[hbt]
\centering
\small
\begin{tabular}{cp{5cm}|p{0.80cm}p{0.80cm}p{0.80cm}|p{0.80cm}p{0.80cm}p{0.80cm}|p{0.80cm}}
\multicolumn{2}{l}{\textbf{Learners}} & \multicolumn{3}{c|}{Infelicitous class} & \multicolumn{3}{c|}{Correct class} & \multicolumn{1}{c}{} \\ \hline
& \multicolumn{1}{c|}{model} & \multicolumn{1}{c}{P} & \multicolumn{1}{c}{R} & \multicolumn{1}{c|}{F1} & \multicolumn{1}{c}{P} & \multicolumn{1}{c}{R} & \multicolumn{1}{c|}{F1} & \multicolumn{1}{c}{acc} \\ \hline
\multirow{3}{*}{\STAB{\rotatebox[origin=c]{90}{$\ge0.8$}}} 
& Gulordava et al. (trained on Reddit) & 0.437 & 0.573 & 0.496 & 0.920 & 0.870 & 0.894 & 0.825 \\
& Google 1B (pre-trained) & 0.500 & 0.686 & 0.578 & 0.946 & 0.889 & 0.917 & 0.861\\
& BERT (pre-trained) & \textbf{0.602} & \textbf{0.736} & \textbf{0.673} & \textbf{0.956} & \textbf{0.911} & \textbf{0.933} & \textbf{0.889} \\
\hline \hline
\multirow{3}{*}{\STAB{\rotatebox[origin=c]{90}{$=1$}}} 
& Gulordava et al. (trained on Reddit) & 0.499 & 0.652 & 0.565 & 0.954 & 0.916 & 0.935 & 0.887 \\
& Google 1B (pre-trained) & 0.523 & 0.720 & 0.606 & 0.970 & 0.932 & 0.950 & 0.912\\
& BERT (pre-trained) & \textbf{0.681} & \textbf{0.859} & \textbf{0.759} & \textbf{0.981} & \textbf{0.949} & \textbf{0.965} & \textbf{0.939} \\
\end{tabular}
\vspace{-0.02in}
\caption{Automatic detection of infelicities in learner data (sentences where annotation disagrees with author usage of IP), with confidence level $\ge0.8$ (top), and with confidence level $=1$ (bottom). 
Baseline accuracy is $0.850$ for the former and $0.887$ for the latter. Best result in a column (for each part) is boldfaced.}
\label{tbl:detection-learners}
\end{table*}

\begin{table*}[hbt]
\centering
\small
\begin{tabular}{cp{5cm}|p{0.80cm}p{0.80cm}p{0.80cm}|p{0.80cm}p{0.80cm}p{0.80cm}|p{0.80cm}}
\multicolumn{2}{l}{\textbf{Advanced L2s}} &  \multicolumn{3}{c|}{Infelicitous class} & \multicolumn{3}{c|}{Correct class} & \multicolumn{1}{c}{} \\ \hline
& \multicolumn{1}{c|}{model} & \multicolumn{1}{c}{P} & \multicolumn{1}{c}{R} & \multicolumn{1}{c|}{F1} & \multicolumn{1}{c}{P} & \multicolumn{1}{c}{R} & \multicolumn{1}{c|}{F1} & \multicolumn{1}{c}{acc} \\ \hline
\multirow{3}{*}{\STAB{\rotatebox[origin=c]{90}{$\ge0.8$}}} 
& Gulordava et al. (trained on Reddit) & 0.274 & 0.583 & 0.373 & 0.959 & 0.863 & 0.908 & 0.840 \\
& Google 1B (pre-trained) & 0.380 & 0.704 & 0.494 & 0.976 & 0.912 & 0.943 & 0.898 \\
& BERT (pre-trained) & \textbf{0.506} & \textbf{0.701} & \textbf{0.585} & \textbf{0.972} & \textbf{0.938} & \textbf{0.955} & \textbf{0.919} \\
\hline \hline
\multirow{3}{*}{\STAB{\rotatebox[origin=c]{90}{$=1$}}} 
& Gulordava et al. (trained on Reddit) & 0.219 & 0.690 & 0.332 & 0.984 & 0.886 & 0.932 & 0.877 \\
& Google 1B (pre-trained) & 0.380 & 0.760 & 0.507 & 0.988 & 0.942 & 0.964 & 0.934 \\
& BERT (pre-trained) & \textbf{0.503} & \textbf{0.790} & \textbf{0.614} & \textbf{0.990} & \textbf{0.964} & \textbf{0.977} & \textbf{0.956} \\
\end{tabular}
\vspace{-0.02in}
\caption{Automatic detection of infelicities in advanced L2 data (sentences where annotation disagrees with author usage of IP), with confidence level $\ge0.8$ (top), and with confidence level $=1$ (bottom). Baseline accuracy is $0.918$ for the former and $0.956$ for the latter. Best result in a column (for each part) is boldfaced.}
\label{tbl:detection-reddit}
\end{table*}

Systematically higher scores obtained for learner utterances (Table~\ref{tbl:detection-learners}), compared to advanced L2s (Table~\ref{tbl:detection-reddit}), imply that the mild infelicities of the latter pose a higher challenge to automatic tools.  That is, not only do advanced L2s show fewer errors, but their errors are likely more subtle and more difficult to detect. The high-confidence setup ($=1.0$) yields results superior to those produced by the lower-confidence setup ($\geq0.8$), further supporting that clear-cut infelicities are more easily captured by an LM.


Returning to our linguistic predictions, the preference of \textit{some-} over \textit{any-} predicted by \textbf{Hypothesis 1} and shown for non-native speakers (\Section{ip-dist}) does not hold for our best-performing LM. We found a roughly equal rate (up to two percent points) of infelicities in model preferences in cases with \textit{some-} vs.\ \textit{any-} gold annotations, showing that the model (unlike non-natives) does not have greater difficulty with \textit{any}- overall.

We also consider the non-uniform difficulty of IPs across various usage cases, predicted by \textbf{Hypothesis 2} and shown for non-natives (\Section{class-dist}). To address this question, we test BERT for infelicitous choices compared to annotators' decisions: That is, for each sentence, we compare the pronoun preferred by the model to the gold annotation. Table~\ref{tbl:infelicities-scorers} presents statistics across usage classes, for learners and advanced L2s (taken from Table~\ref{tbl:infelicities-dist}), as well as for BERT. The top panel refers to learner 
data; the bottom panel, to advanced L2 data.
While (expectedly) outperforming the two non-native populations, the model exhibits similar distributional patterns, with more infelicities in the CD and QU classes.
The model also has a higher number of infelicities in the CP class for learners; again, we note the small sample of data in this class, entailing a need for further investigation of this particular pattern.
The model results here pose intriguing questions for future work regarding the nature of challenges faced by automatic neural methods, and their potential analogues to those of humans.

\begin{table}[hbt]
\centering
\resizebox{\columnwidth}{!}{
\begin{tabular}{l|rggrr}
& DN & QU & CD & CP & MIXED \\
\midrule[0.01mm]
learners & 8.1 & 24.0 & 12.4 & 8.9 & 8.7 \\
BERT &  0.8 & 6.1 & 3.6 & 4.0 & 2.2 \\
\midrule[0.01mm]
advanced L2s & 5.3 & 7.1 & 9.1 & 5.1 & 5.7 \\
BERT &  1.3 &  2.5 &  2.7 &  1.6 &  1.5 \\
\end{tabular}
}
\vspace{-0.02in}
\caption{{Distribution of \% of infelicities (difference from gold annotation) across classes for humans and for BERT on the corresponding data.}}
\vspace{-0.02in}
\label{tbl:infelicities-scorers}
\end{table}

\section{Related Work}
\label{sec:error-detection}
Computational approaches to grammatical error correction (GEC) in learners' productions has been a prolific field of research in recent years. A standard approach to dealing with grammar and spelling errors makes use of a machine-learning classification paradigm; a comprehensive survey of these methods can be found in  \citet{ng2014conll}. Recent advances in the field of GEC were achieved by using neural models \cite{yuan2016grammatical, ji2017nested, sakaguchi2017grammatical, lo2018cool}. Most studies used a \textit{supervised} setup for selecting a correct choice (e.g., a preposition) out of a set of multiple alternatives, rendering our experimental setup not directly comparable.

Another line of work has assessed the capability of neural LMs to capture errors stemming from violation of syntax-sensitive dependencies \cite{linzen2016assessing, gulordava2018colorless, marvin2018targeted}.  The recent BERT model \cite{devlin2018bert} has been shown to be highly effective for detection of syntactic anomalies stemming from subject-verb disagreement \cite{goldberg2019assessing}.

Most research on L2 error correction focuses on function words, such as prepositions and determiners. Very little work has been done on detecting and correcting incorrect usage of content words. Most has been focused on the felicity of word combinations, such as identifying disfluencies stemming from L1 paraphrases \citep[e.g., \textit{eat medicine} or \textit{look movies},][]{Brooke2011,dahlmeier2011correcting}, or using models of compositionality to detect semantically deviant pairs \citep[\textit{residential steak},][]{vecchi2011linear} or infelicitous collocations \citep[?\textit{big importance} vs.\ \textit{great importance},][]{kochmar2013capturing}. A shared task on automatic evaluation of scientific writing \citep[][]{daudaravicius2016report} addressed automatic detection of a variety of grammatical errors (e.g., misuse of an article or punctuation) and lexical infelicities (e.g., phrasing choices stemming from style requirements of the genre) in scientific papers, edited by a professional company.

While most closely related to the field of semantic error detection, our work deals with subtle linguistic choices that shape the ultimate attainment of L2 in non-native speakers. Compared to grammatical and semantic anomalies explored in previous work, the choice of indefinite pronoun is often guided by implicit contextual clues that are not necessarily reflected in superficial collocational patterns, thereby posing a higher challenge for automatic techniques.

\section{Conclusion}

We develop and evaluate linguistic hypotheses on the difficulties for second language learners of the atypical system of English indefinite pronouns. We find that the tangled relation between \textit{some-} and \textit{any-} pronouns pose challenges that are evident in the productions of both learners and advanced L2 speakers.  This work thus demonstrates the promise of extending computational approaches for error-detection in L2 productions to more subtle semantic usages.  Moreover, our results reveal the challenges that these subtleties can pose for even advanced non-native speakers.

Much research in second language acquisition establishes \textit{native language transfer} as one of the major factors that shape productions of non-native speakers. While the work here addresses \textit{universal} (i.e., native-language independent) challenges posed to L2 speakers, a plausible assumption is that mastery of English IPs is also affected by the proximity of the analogous system in a speaker's L1. We leave this direction for future research.

We also evaluate here the ability of language models to detect the errors arising in the use of English indefinite pronouns in L2 productions.  Not surprisingly, we find that the more clearcut errors exhibited by learners are easier to automatically identify than the potentially more subtle errors that arise with advanced L2 speakers.  The best performing language model shows a varying match to human patterns of difficulty, raising issues for further research regarding the factors that influence difficulty for both humans and language models.

The practical impact of this work will be in facilitating the development of educational applications for L2 English speakers at various levels of proficiency. At present, most error correction and detection tools focus on explicit spelling or grammar errors.  Enriching these tools with the ability to capture subtle semantic infelicities in the usage of IPs would advance the current state of the art in educational applications for language learners.

\section*{Acknowledgments}
This research is supported by an NSERC Discovery Grant RGPIN-2017-06506 to Suzanne Stevenson. We are thankful to Paola Merlo for her insight and advice. We are also grateful to our anonymous reviewers for their constructive feedback.

\bibliographystyle{acl_natbib}
\bibliography{main}

\end{document}